\documentclass{article}

\PassOptionsToPackage{numbers, compress}{natbib}
\usepackage[final]{nips_2017}

\usepackage[utf8]{inputenc} 
\usepackage[T1]{fontenc}    
\usepackage{hyperref}       
\usepackage{url}            
\usepackage{booktabs}       
\usepackage{amsfonts}       
\usepackage{nicefrac}       
\usepackage{microtype}      

\usepackage{fancyhdr}
\usepackage{datetime}
\usepackage{amsmath}
\usepackage{float}

\usepackage{graphicx}
\usepackage{subcaption}
\graphicspath{ {figures/} }

\title{\textit{Dex}: Incremental Learning for Complex Environments in Deep Reinforcement Learning}

\author{
  Nick Erickson, Qi Zhao \\
  Department of Science and Engineering\\
  University of Minnesota, Twin Cities\\
  Minneapolis, MN 55455\\
  \texttt{eric3068@umn.edu, qzhao@cs.umn.edu} \\
}

\begin{document}

\maketitle

\begin{abstract}

This paper introduces \textit{Dex}, a reinforcement learning environment toolkit specialized for training and evaluation of continual learning methods as well as general reinforcement learning problems. We also present the novel continual learning method of incremental learning, where a challenging environment is solved using optimal weight initialization learned from first solving a similar easier environment.  We show that incremental learning can produce vastly superior results than standard methods by providing a strong baseline method across ten \textit{Dex} environments. We finally develop a saliency method for qualitative analysis of reinforcement learning, which shows the impact incremental learning has on network attention.

\end{abstract}

\section{Introduction}

Complex environments such as \textit{Go}, \textit{Starcraft}, and many modern video-games present profound challenges in deep reinforcement learning that have yet to be solved. They often require long, precise sequences of actions and domain knowledge in order to obtain reward, and have yet to be learned from random weight initialization. Solutions to these problems would mark a significant breakthrough on the path to artificial general intelligence.

Recent works in reinforcement learning have shown that environments such as Atari games \cite{ALE} can be learned from pixel input to superhuman expertise \cite{mnih2015human}. The agents start with randomly initialized weights, and learn largely from trial and error, relying on a reward signal to indicate performance. Despite these successes, complex games, including those where rewards are sparse such as \textit{Montezuma's Revenge}, have been notoriously difficult to learn. While methods such as intrinsic motivation \cite{intrinsic} have been used to partially overcome these challenges, we suspect this becomes intractable as complexity increases. Additionally, as environments become more complex, they will become more expensive to simulate. This poses a significant problem, since many Atari games already require upwards of 100 million steps using state-of-the-art algorithms, representing days of training on a single machine.

Thus, it appears likely that complex environments will become too costly to learn from randomly initialized weights, due both to the increased simulation cost as well as the inherent difficulty of the task. Therefore, some form of prior information must be given to the agent. This can be seen with \textit{AlphaGo} \cite{AlphaGo}, where the agent never learned to play the game without first using supervised learning on human games. While supervised learning certainly has been shown to aid reinforcement learning, it is very costly to obtain sufficient samples and requires the environment to be a task humans can play with reasonable skill, and is therefore impractical for a wide variety of important reinforcement learning problems.


In this paper we introduce \textit{Dex}, the first continual reinforcement learning toolkit for training and evaluating continual learning methods. We present and demonstrate a novel continual learning method we call incremental learning to solve complex environments. In incremental learning, environments are framed as a task to be learned by an agent. This task can be split into a series of subtasks that are solved simultaneously.

Similar to how natural language processing and object detection are subtasks of neural image caption generation \cite{ImageCaption}, reinforcement learning environments also have subtasks relevant to a given environment. These subtasks often include player detection, player control, obstacle detection, enemy detection, and player-object interaction, to name a few. These subtasks are common to many environments, but they are often sufficiently different in function and representation that reinforcement learning algorithms fail to generalize them across environments, such as in Atari. These critical subtasks are what expert humans utilize to quickly learn in new environments that share subtasks with previously learned environments, and are a reason for humans superior data efficiency in learning complex tasks.

In the case of deliberately similar environments, we can construct the subtasks such that they are similar in function and representation that an agent trained on the first environment can accelerate learning on the second environment due to its preconstructed subtask representations, thus partially avoiding the more complex environment's increased simulation cost and inherent learning difficulty.

\section{Related work}

Transfer learning \cite{transfer_survey} is the method of utilizing data from one domain to enhance learning of another domain. While sharing significant similarities to continual learning, transfer learning is applicable across all machine learning domains, rather than being confined to reinforcement learning. For example, it has had significant use with using networks trained on ImageNet \cite{imagenet} to accelerate or enhance learning and classification accuracy by finetuning in a variety of vision tasks \cite{imagenet_transfer}. 

While the concept of continual learning, where an agent learns from a variety of experiences to enhance future learning, has been defined for some time \cite{child}, it has remained largely untapped by recent powerful algorithms that are best fit to benefit from its effects. Recent work has been done with \textit{Progressive Neural Networks} \cite{progressive_neural_networks}, where transfer learning was used to apply positive transfer in a variety of reinforcement learning domains. However, our method differs in that it does not add additional parameters for each environment or use lateral connections to features that result in increased memory space and training time.

Recent work related to subtask utilization comes from \citet{catastrophic_full}, which shows that expertise can be maintained on multiple environments that have not been experienced for a long time through elastic weight consolidation (EWC). Viable weights were found that simultaneously achieve expertise in a variety of Atari games. Incremental learning similarly trains on multiple environments, but with the goal of achieving enhanced expertise in a single environment, rather than expertise in all environments. We leave it to future work to overcome this limitation.

\section{\textit{Dex}}

\textit{Dex} is a novel deep reinforcement learning toolkit for training and evaluating agents on continual learning methods. \textit{Dex} acts as a wrapper to the game \textit{Open Hexagon} \cite{openHexagon}, sending screen pixel information, reward information and performing actions via an OpenAI Gym like API \cite{OpenAIGym}. \textit{Dex} contains hundreds of levels, each acting as their own environment. These environments are collected into groups of similar environments for the task of continual learning. \textit{Dex} environments vary greatly in difficulty, ranging from very simple levels where agents achieve superhuman performance in less that 4 minutes, to levels we consider far more complex than any previously learned environments.

\begin{figure}[htbp]
\centering
\begin{subfigure}{.16\textwidth}
  \centering
  \includegraphics[width=.95\linewidth]{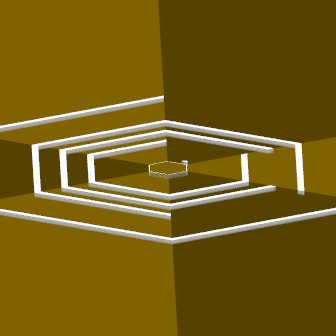}
  \caption{Distortion}
  \label{fig:sfig0}
\end{subfigure}%
\begin{subfigure}{.16\textwidth}
  \centering
  \includegraphics[width=.95\linewidth]{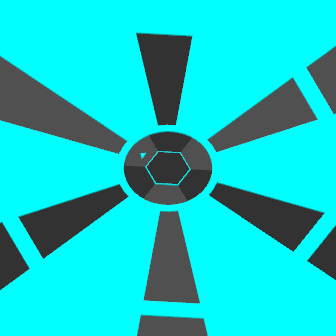}
  \caption{Lanes}
  \label{fig:sfig1}
\end{subfigure}
\begin{subfigure}{.16\textwidth}
  \centering
  \includegraphics[width=.95\linewidth]{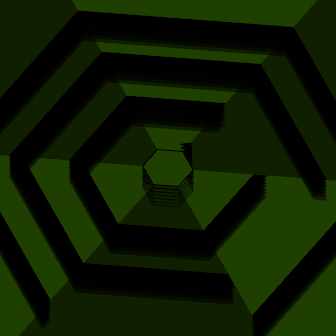}
  \caption{Reversal}
  \label{fig:sfig2}
\end{subfigure}
\begin{subfigure}{.16\textwidth}
  \centering
  \includegraphics[width=.95\linewidth]{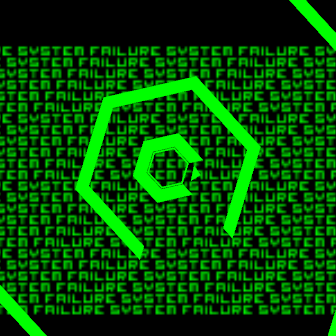}
  \caption{System}
  \label{fig:sfig3}
\end{subfigure}
\begin{subfigure}{.16\textwidth}
  \centering
  \includegraphics[width=.95\linewidth]{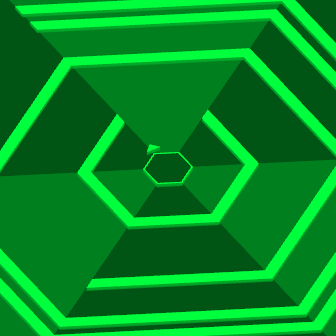}
  \caption{Desync}
  \label{fig:sfig4}
\end{subfigure}
\begin{subfigure}{.16\textwidth}
  \centering
  \includegraphics[width=.95\linewidth]{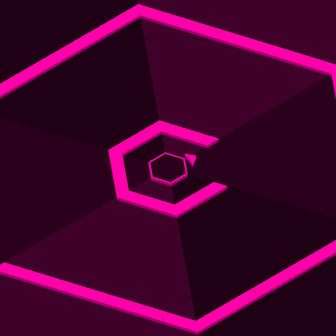}
  \caption{Stutter}
  \label{fig:sfig5}
\end{subfigure}
\begin{subfigure}{.16\textwidth}
  \centering
  \includegraphics[width=.95\linewidth]{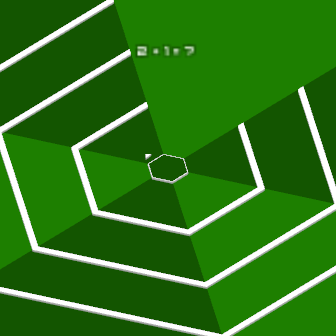}
  \caption{Arithmetic}
  \label{fig:sfig6}
\end{subfigure}
\begin{subfigure}{.16\textwidth}
  \centering
  \includegraphics[width=.95\linewidth]{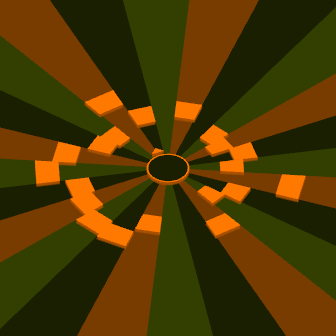}
  \caption{Stretch}
  \label{fig:sfig7}
\end{subfigure}
\begin{subfigure}{.16\textwidth}
  \centering
  \includegraphics[width=.95\linewidth]{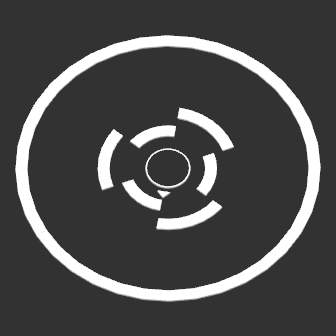}
  \caption{Open}
  \label{fig:sfig8}
\end{subfigure}
\begin{subfigure}{.16\textwidth}
  \centering
  \includegraphics[width=.95\linewidth]{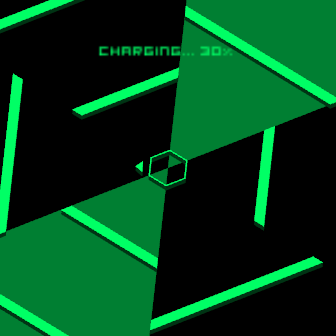}
  \caption{Overcharge}
  \label{fig:sfig9}
\end{subfigure}
\begin{subfigure}{.16\textwidth}
  \centering
  \includegraphics[width=.95\linewidth]{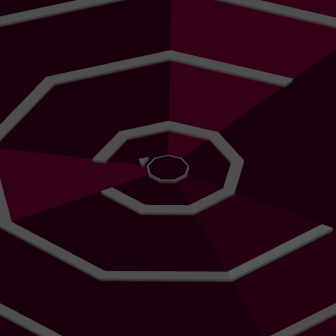}
  \caption{Enneagon}
  \label{fig:sfig10}
\end{subfigure}
\begin{subfigure}{.16\textwidth}
  \centering
  \includegraphics[width=.95\linewidth]{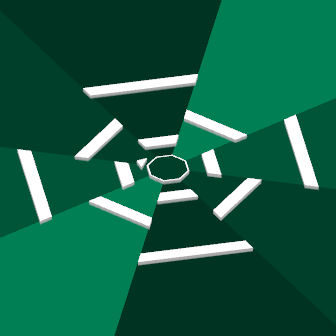}
  \caption{Blink}
  \label{fig:sfig11}
\end{subfigure}
\begin{subfigure}{.16\textwidth}
  \centering
  \includegraphics[width=.95\linewidth]{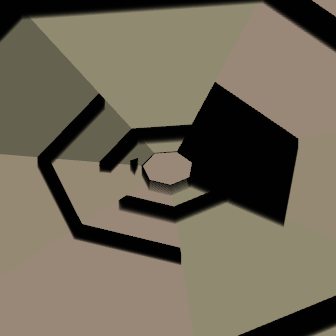}
  \caption{Tunnel Town}
  \label{fig:sfig12}
\end{subfigure}
\begin{subfigure}{.16\textwidth}
  \centering
  \includegraphics[width=.95\linewidth]{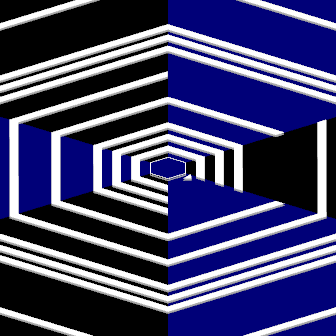}
  \caption{Roulette}
  \label{fig:sfig13}
\end{subfigure}
\begin{subfigure}{.16\textwidth}
  \centering
  \includegraphics[width=.95\linewidth]{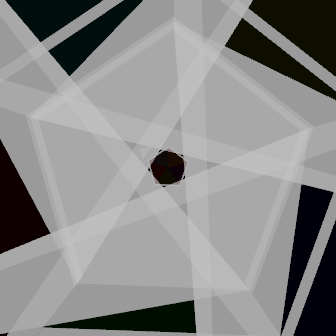}
  \caption{Echnoderm}
  \label{fig:sfig14}
\end{subfigure}
\begin{subfigure}{.16\textwidth}
  \centering
  \includegraphics[width=.95\linewidth]{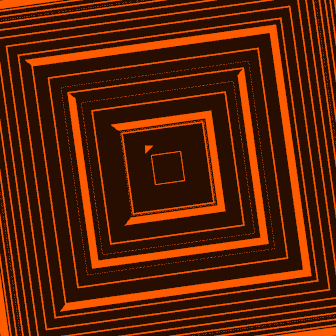}
  \caption{Transmission}
  \label{fig:sfig15}
\end{subfigure}
\begin{subfigure}{.16\textwidth}
  \centering
  \includegraphics[width=.95\linewidth]{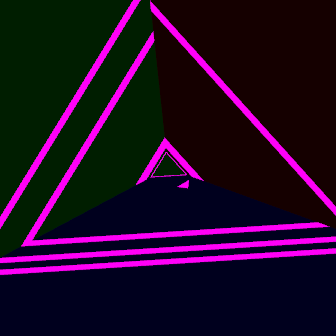}
  \caption{Triptych}
  \label{fig:sfig16}
\end{subfigure}
\begin{subfigure}{.16\textwidth}
  \centering
  \includegraphics[width=.95\linewidth]{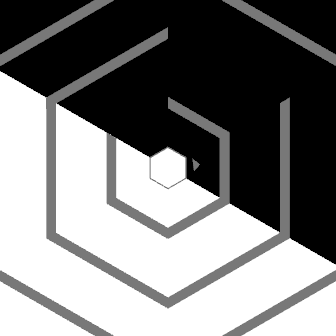}
  \caption{Duality}
  \label{fig:sfig17}
\end{subfigure}
\begin{subfigure}{.16\textwidth}
  \centering
  \includegraphics[width=.95\linewidth]{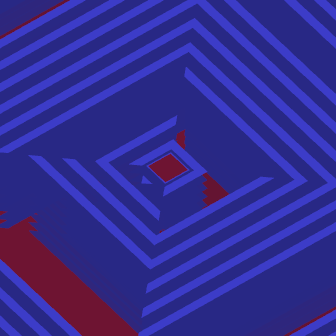}
  \caption{World Battle}
  \label{fig:sfig18}
\end{subfigure}
\begin{subfigure}{.16\textwidth}
  \centering
  \includegraphics[width=.95\linewidth]{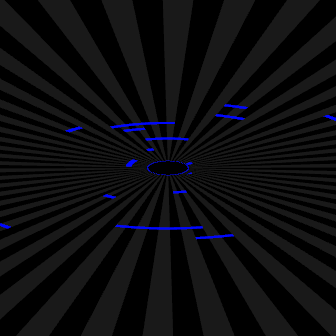}
  \caption{Universe}
  \label{fig:sfig19}
\end{subfigure}
\begin{subfigure}{.16\textwidth}
  \centering
  \includegraphics[width=.95\linewidth]{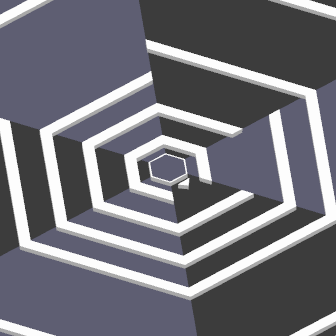}
  \caption{Apeirogon}
  \label{fig:sfig20}
\end{subfigure}
\begin{subfigure}{.16\textwidth}
  \centering
  \includegraphics[width=.95\linewidth]{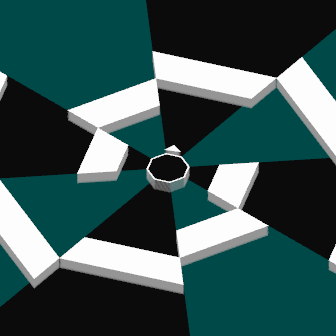}
  \caption{Euclidean}
  \label{fig:sfig21}
\end{subfigure}
\begin{subfigure}{.16\textwidth}
  \centering
  \includegraphics[width=.95\linewidth]{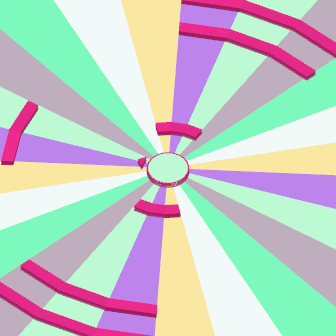}
  \caption{Pi}
  \label{fig:sfig22}
\end{subfigure}
\begin{subfigure}{.16\textwidth}
  \centering
  \includegraphics[width=.95\linewidth]{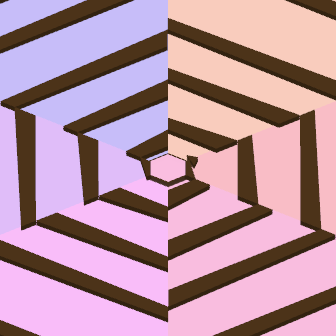}
  \caption{Golden Ratio}
  \label{fig:sfig23}
\end{subfigure}
\caption{\textit{Dex} environments. The small triangle is the player that must be rotated around the center to avoid incoming walls. Many of these environments incorporate various distortion effects that are not evident in screenshots. Reversal periodically flips game controls, and some environments even add additional actions, such as in Arithmetic, which requires the agent to correctly solve various math equations during the level through the numpad.}
\label{fig:hex}
\end{figure}

Refer to videos available at \href{https://github.com/innixma/dex}{\texttt{github.com/innixma/dex}} of the \textit{Dex} environments shown in \textit{Figure \ref{fig:hex}}, as screenshots do not capture the environment complexity.

\textit{Open Hexagon} is a game involving navigating a triangle around a center figure to avoid incoming randomly generated walls. Screenshots of various environments from the game are shown in \textit{Figure \ref{fig:hex}}. The game progresses regardless of player action, and thus the player must react to the environment in real-time. If the player contacts a wall the game is over. At each point in the game, a player has only three choices for actions: move left, right, or stay put. It is a game of survival, with the score and thus total reward being the survival time in seconds. \textit{Open Hexagon} contains hundreds of levels drastically ranging in difficulty, yet they each contain many similar core subtasks. This makes \textit{Open Hexagon} an ideal platform for testing continual learning methods.

The \textit{Dex} toolkit along with its source code is available at \href{https://github.com/innixma/dex}{\texttt{github.com/innixma/dex}}.

\section{Incremental learning}

The novel continual learning method of incremental learning is defined as follows. 

In the formal case, an agent must learn from a series of $n$ environments $\mathcal{E}$, each with identical legal game actions $\mathcal{A} = \{1, ..., K\}$. Note that any series of environments can be made to have the same action space by considering $\mathcal{A}$ to be the superset of all possible actions in each game, with new actions performing identically to no action, assuming no action is within the action space.

Each environment $\mathcal{E}_i$ has a corresponding cost per step $c_i > 0$ and a step count $s_i \geq 0$, indicating the number of steps taken in that environment. Typically, more complex environments will have a higher cost per step. A resource maximum $M$ is given, which indicates the total amount of data that can be gathered, shown in the following inequality:

\[M \geq \sum_{i=1}^{n} c_i s_i\]

The problem is to maximize the mean total reward $R$ of the agent in the goal environment, $\mathcal{E}_n$, while maintaining the above inequality. Steps can be taken in any order from the $n$ environments, and there is no assumed correlation between the environments beyond their data and action dimensions. While it may appear an optimal solution to only examine and gather data from $\mathcal{E}_n$, as has been done for virtually all reinforcement learning algorithms in the past, this is not always the case. For example, if $\mathcal{E}_{n-1}$ and  $\mathcal{E}_n$ are highly correlated, and $c_{n-1} << c_{n}$, then training with $\mathcal{E}_{n-1}$ may be superior due to its lesser cost. Additionally, an environment $\mathcal{E}_{n-1}$ may contain important aspects of $\mathcal{E}_{n}$, while avoiding state spaces and rewards that are not useful for training, potentially allowing training on $\mathcal{E}_{n-1}$ to be optimal, even if $c_{n-1} > c_{n}$.

By taking environments $\mathcal{E}$ to represent all real environments with their respective costs, the solution to this problem corresponds to the globally optimal sequence of training steps to achieve maximum performance with a finite amount of computational resources for a given algorithm. It therefore necessarily contains the solution of achieving artificial general intelligence with minimal resources. Unfortunately, this has several drawbacks. Most importantly, the selection of the optimal environments is not obvious, and their order even less so.

While this formal definition is useful to define incremental learning, the following simplified version will be what this paper focuses on. In the simplified case, all variables are the same, except that steps must be taken in environments sequentially, without going back to previous environments. Thus, once a step has been taken in $\mathcal{E}_{i}$, no steps may be taken in future environments $\mathcal{E}_{j}$ where $j < i$. Furthermore, it is assumed that the environments are correlated, where environment $\mathcal{E}_{i-1}$ contains a subset of subtasks in environment $\mathcal{E}_{i}$, with environment $\mathcal{E}_i$ being typically harder than environment $\mathcal{E}_{i-1}$.

The intuition behind this simplified case is that simple environments are both cheap to simulate and easy to learn, and that the features and strategies learned from the simple environment could transfer to a more difficult correlated environment. This process can be done repeatedly, producing a compounding acceleration of learning as additional useful incremental environments are added. Thus, the general process of incremental environment selection is to use easier subsets of the goal environment. Furthermore, this means that every environment $\mathcal{E}_{i}$ in a simplified incremental learning problem can be seen as the goal environment in an easier problem containing $\mathcal{E}_{1:i}$.

\section{Baseline learning algorithm and model architecture}

To analyse the effectiveness of incremental learning, our agents learned environments from \textit{Dex}.
For training agents, we use Asynchronous Advantage Actor-Critic (A3C) framework introduced in \cite{a3c}, coupled with the network architecture described below.

\textbf{ConvNet implementation details.} All \textit{Dex} experiments were learned by a network with the following architecture. The network takes as input a series of 2 consecutive $42\times42$ grayscale images of a game state. It is then processed through a series of 3 convolutional layers, of stride $1\times1$ and size $5\times5$, $4\times4$, and $3\times3$ with filter counts of 32, 64, and 64 respectively. Between each pair of convolutional layers is a 2 dimensional maxpooling layer, of size $2\times2$. The intermediate output is then flattened and processed by a dense layer of 512 nodes. The output layers are identical to those specified for A3C. All convolutional layers and maxpooling layers are zero-padded. This results in an network with approximately 4,000,000 parameters. All intermediate layers are followed by rectified linear units (ReLU) \cite{relu}, and Adam is used for the optimizer, with a learning rate of $10^{-3}$. Learning rate is not changed over the course of training as opposed to \citet{a3c}, but instead stays constant. A gamma of 0.99 is used, along with \textit{n-step} reward \cite{nstep} with $n=4$ , which is used to speed up value propagation in the network, at a cost of minor instability in the learning. Training is done with batches of 128 samples. The architecture was created using TensorFlow 1.1.0 \cite{tensorflow}, and Keras 2.0.3.

Since \textit{Dex} runs in real time, a slightly altered method of learning was utilized to avoid inconsistent time between steps. Our implemented A3C was modified to work in an offline manner, with experience replay as done in Deep-Q Networks \cite{dqn}. This is a naive approximation to the more complex ACER algorithm \cite{acer}. While this does destabilize the algorithm, which bases its computations on data being representative of the network in its current state, the agents are still able to learn the environments and significantly outperform Double Deep-Q Networks \cite{ddqn}, thus serving as a reasonable baseline.

\section{Experiments}

So far, we have performed experiments on ten different environments in \textit{Dex} for incremental learning. These ten environments are split into two incremental learning sets of three and seven environments. The first set will be referred to as $a$, and deals with increasingly complex patterns. The second set will be referred to as $b$, and deals with increasingly complex task representation. The results show that incremental learning can have a significant positive impact on learning speed and task performance, but can also lead to bad initializations when trained on overly simplistic environments.

Code to reproduce the experiments in this paper will be released at a future date.

\subsection{Setup}

In the following experiments, we naively assume that achieving near optimal performance in $\mathcal{E}_{i}$ with minimal resources requires as a prerequisite learning $\mathcal{E}_{i-1}$ with minimal resources. This proceeds downward to the base case of $\mathcal{E}_{1}$, which can be considered a trivially solvable environment from random weight initialization. This assumption is used to simplify training. Furthermore, due to the exploratory nature of the baseline experiments, the costs per step are ignored, as we seek only to show that positive feature transfer is occurring, rather than optimizing the feature transfer itself, which we leave for future work.

All environments are learned with identical architecture, algorithm and hyperparameters. We act and gather data on the environments 50 times per second. We use an $\epsilon$-greedy policy with $\epsilon = 0.05$, and a replay memory size of 40,000. Replay memory is initialized by a random agent. For all experiments, agents are trained with 75 batches after each episode. Total reward is equal to the number of seconds survived in the environment each episode. Mean reward is calculated from an hour of testing weights without training. Each episode scales in difficulty indefinitely the longer the agent survives.

\subsection{Models}

We evaluate four different types of models in this experiment. The first is a random agent for comparison. The second is the baseline, which is the standard reinforcement learning training method. To establish the baseline, each environment $\mathcal{E}_{i}$ is trained on from random initialization for one hour, equivalent to roughly 150,000 training steps. The weights which achieve the maximum total reward in a single episode are selected as the output weights from the training, called $w_{i}$.

A third model, which we shall call \textit{initial}, is the initial weights to the incremental learning method before continued training. Thus, for environment $\mathcal{E}_{i}$ this model uses weights $w_{i-1}$. This is used to measure the correlation between the two environments. We would expect uncorrelated environments to result in near random agent reward for \textit{initial}.

To establish the incremental learning agents, for each environment $\mathcal{E}_{i}$ we take the weights $w_{i-1}$ outputted by the baseline, using them as the initial weights to training on $\mathcal{E}_{i}$ for an additional hour. The weights outputted by this method we call $w'_{i}$.

\subsection{Results and Observations}

The results can be found in \textit{Table \ref{table:b}} and \textit{Table \ref{table:a}}.

\begin{table}[!htbp]
  \centering
  \caption{Set $b$ rewards}
  \label{table:b}
  \begin{tabular}{ccccc}
    \toprule
    Max \\
    \midrule
    $\mathcal{E}$     & Random & Initial & Baseline & Incremental\\
    \midrule
    $b_1$     & 31.43 & --- & 425.92   & --- \\
    $b_2$     & 8.23 & 120.77 & 259.80   & 280.59\\
    $b_3$     & 7.73 & 58.30 & 132.31   & 221.96 \\
    $b_4$     & 8.27 & 35.10 & 35.66   & 105.48  \\
    $b_5$     & 8.79 & 19.76 & 52.81   & 41.57  \\
    $b_6$     & 9.33 & 11.05 & 10.79   & 13.11  \\
    $b_7$     & 9.97 & 7.13 & 9.59   & 14.54  \\ 
    \bottomrule
    \\
    Mean \\
    \midrule
    $\mathcal{E}$     & Random & Initial & Baseline & Incremental \\
    \midrule
    $b_1$     & 9.10 & --- & 169.54   & --- \\
    $b_2$     & 2.23 & 36.05 & 92.24   & 85.52 \\ 
    $b_3$     & 2.43 & 8.22 & 41.17   & 66.32 \\
    $b_4$     & 2.17 & 5.30 & 8.11   & 26.75  \\
    $b_5$     & 1.88 & 4.17 & 12.23   & 11.61 \\
    $b_6$     & 2.17 & 2.89 & 2.15   & 2.25 \\
    $b_7$     & 2.30 & 2.34 & 2.08   & 2.58 \\ 
    \bottomrule
    \\
  \end{tabular}
  \caption*{This table shows the max and mean reward for an agent on the given environment $\mathcal{E}$ with a given training method, as described in the experimental setup. Here we observe that incremental learning provided superior results to the baseline in nearly all environments, particularly $b_3$ and $b_4$.}
\end{table}

\begin{table}[!htbp]
  \centering
  \caption{Set $a$ rewards}
  \label{table:a}
  \begin{tabular}{ccccc}
    \toprule
    Max \\
    \midrule
    $\mathcal{E}$     & Random & Initial & Baseline & Incremental \\
    \midrule
    $a_1$     & 40.73 & --- & 771.67   & ---     \\
    $a_2$     & 19.65 & 53.37 & 445.85   & 86.57     \\
    $a_3$     & 10.85 & 15.69 & 49.50    & 59.10  \\
    \bottomrule
    \\
    Mean \\
    \midrule
    $\mathcal{E}$     & Random & Initial & Baseline & Incremental \\
    \midrule
    $a_1$     & 8.93 & --- & 717.93   & ---    \\
    $a_2$     & 7.46 & 18.34 & 87.06   & 18.31      \\
    $a_3$     & 6.01 & 7.32 & 9.81    & 13.52  \\
    \bottomrule
    \\
  \end{tabular}
  \caption*{This table shows the max and mean reward for an agent on the given environment $\mathcal{E}$ with a given training method, as described in the experimental setup. Here we observe that incremental learning provided inferior results to the baseline in the $a_2$ environment, likely due to overfitting.}
\end{table}

As can be seen in \textit{Table \ref{table:b}}, incremental learning provided superior or roughly equivalent results on every environment in set $b$, with some environments such as $b_3$ and $b_4$ experiencing substantial increases in maximum reward, with the incrementally learned $b_4$ achieving nearly triple the baseline in both maximum and mean reward. However, on the harder environments, there was not significant improvement seen. This is likely because the earlier environments were not sufficiently learned along with the fact that the tasks were generally too difficult to learn in one hour of training, indicated by the near random performance of $b_6$ and $b_7$ on both the baseline and incremental methods.

The other set, results shown in \textit{Table \ref{table:a}} for set $a$, show that incremental learning was harmful in the case of $a_2$. This is likely due to the difference in the wall patterns of $a_1$ and $a_2$. In $a_1$, a single wall is on the screen at a time, requiring a simple avoidance. In $a_2$, up to four separate walls occupy the screen at once, requiring a more complex method involving future planning and understanding of which walls are more important in a given state. We suspect that learning $a_1$ leads the agent to overtrain on a variety of weights, hindering future learning. This indicates that certain environments may be too simple to include in incremental learning, as $a_1$ can be learned to a reward of over 700 in less than four minutes.

Additionally, the \textit{initial} model shows that the environments are correlated, with generally far superior performance than random, despite never training on the environment explicitly. In the case of $b_4$, it is nearly equal to the baseline in the maximum metric, indicating significant correlation. This is likely a reason for the greatly enhanced performance of incremental learning over the baseline in $b_4$.


Note that these experimental results are strictly a simple baseline for both \textit{Dex} and incremental learning, and suffer from instability due to the short timeline of training and lack of repeated experiments. This means that agents do not necessarily consistently improve throughout training, but rather may quickly improve to maximum performance followed by decreased performance for the remainder of training. We leave more comprehensive experiments to future work.


\section{Visualization}

\begin{figure}[H]
\center{
	\includegraphics[width=0.42 \linewidth]{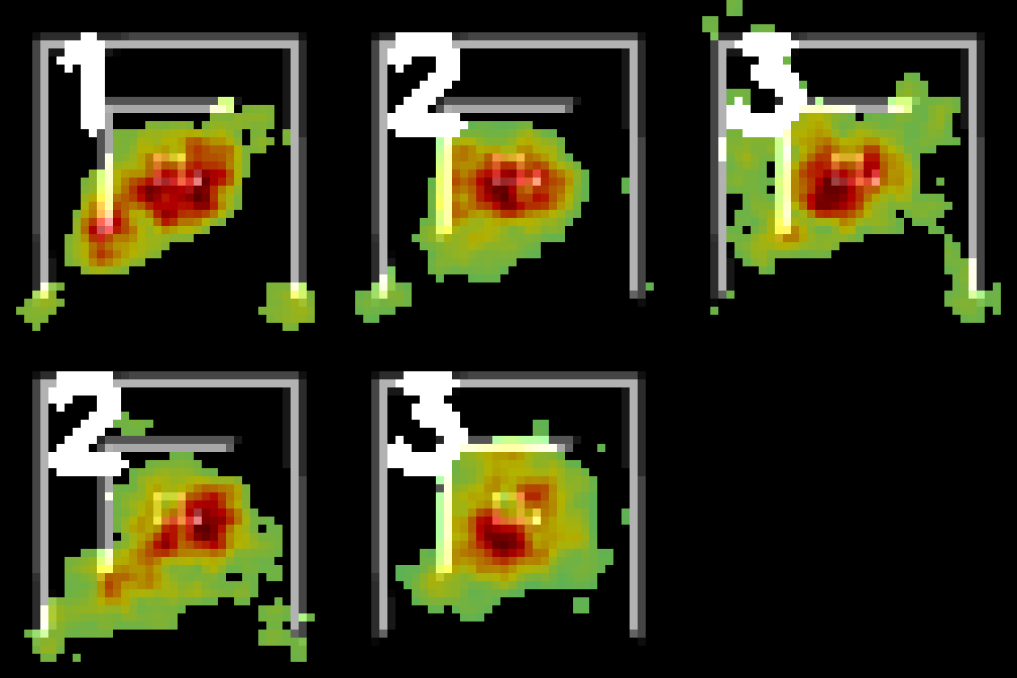}}
\caption{Saliency mappings of a state from environment $a_3$, with weights from set $a$. The first row consists of the trained baseline weights $w_1$ to $w_3$. The second row consists of the incrementally learned weights $w'_2$ to $w'_3$.}
\label{fig:visb}
\end{figure}

\begin{figure}[H]
\center{
	\includegraphics[width=1 \linewidth]{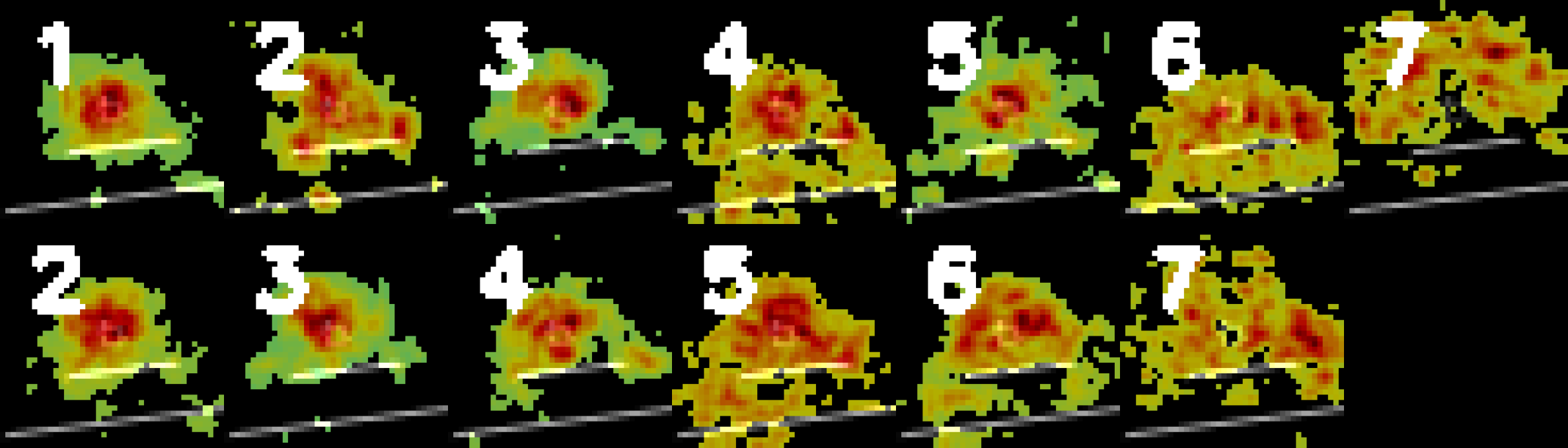}}
\caption{Saliency mappings of a state from environment $b_3$, with weights from set $b$. The first row consists of the trained baseline weights $w_1$ to $w_7$. The second row consists of the incrementally learned weights $w'_2$ to $w'_7$.}
\label{fig:visr}
\end{figure}

To qualitatively analyze the effects of incremental learning on a networks weights, we develop a saliency visualization method based on \citet{saliency} for reinforcement learning. Heatmaps are generated with a given networks weights and an input image. The method for gathering the heatmaps is identical to \citet{saliency}, and thus equations and derivations shall not be repeated in this paper. The most likely action the network will take at a given frame is used for the action to minimize through the gradient. This is a difference from the supervised learning case, where the ground truth is used. In reinforcement learning, the ground truth is not known, and thus must be inferred, as done in \citet{dueling}. Results of the visualization on the trained weights of environment sets $a$ and $b$ can be seen in \textit{Figure \ref{fig:visb}} and \textit{Figure \ref{fig:visr}}.

As can be expected, saliency of a well performing trained agent will focus on the player location, being very sensitive to changes in a small region surrounding the player. This intuition is confirmed in the first row of \textit{Figure \ref{fig:visr}}. The agents trained on the easier environments, such as $b_1$, $b_2$, and $b_3$, focus attention on the player and nearby threats. Interestingly, as the trained environment becomes more complex, the agent appears less developed, and increasingly focuses on irrelevant locations in the state, most prominent in the agent trained on $b_7$, which explicitly avoids attention on the player and nearby threats, such as the walls. This indicates that the agent did not learn its environment sufficiently in the time it was given, due to the complexity of its training environment. This is further confirmed through the experimental results of $b_6$ and $b_7$, which were near random.

Additionally, the saliency mappings show the correlation between the environments, as indicated by the similarity in saliency in $w_1$ and $w_2$ to $w_3$ in both \textit{Figure \ref{fig:visb}} and \textit{Figure \ref{fig:visr}}, despite never explicitly being trained on the environment that the state in the visualization is from.

Interestingly, the saliency mappings of the incrementally learned agents more closely resemble the mappings of the weights it was initialized to than the weights its environment learned without incremental learning. This suggests that a significant amount of the initialized weights are retained after incrementally learning.

Further visualizations are included as a video of realtime saliency of an agent episode at \href{https://github.com/innixma/dex}{\texttt{github.com/innixma/dex}}.

\section{Future work}

We hope to expand the analysis done in this paper by investigating incremental learning chains involving more than two environments. This will more effectively explore the effects of incremental learning in complex problems. We also wish to extend the training time of experiments to allow for more complex environments such as those shown in \textit{Figure \ref{fig:hex}} to be learned, as well as to compare our method to \textit{Progressive Neural Networks} \cite{progressive_neural_networks}. Incremental learning could be expanded to function as a feature extractor in reinforcement learning. This would allow a more contained action space for incremental learning tasks of varying domains, and is similar to the work done in \citet{progressive_neural_networks}. 

Additionally, training could be done separately on multiple environments and then combined to learn an environment that shares subproblems with all the previous environments. This would be a natural merging of incremental learning and elastic weight consolidation \cite{catastrophic_full}. It may also provide a synergistic effect to algorithms such as \textit{UNREAL} \cite{unreal} which rely on auxiliary tasks. These tasks could act to maximize network utilization across multiple environments, potentially leading the network to better generalizations through incremental learning.

Finally, environments such as \textit{Montezuma's Revenge} and Labyrinth \cite{a3c} which require exploration with sparse rewards are a natural expansion to the application of incremental learning. Developing incremental exploration environments could result in far superior performance on exploration tasks. 

\section{Conclusion}

The ability to learn and transfer knowledge across domains is essential to the advancement of agents that can solve complex tasks. This paper introduced the continual learning toolkit \textit{Dex} for training and evaluation of continual learning methods. We proposed the method of incremental learning for deep reinforcement learning, and demonstrated its ability to accelerate learning and produce drastically superior results to standard training methods in multiple \textit{Dex} environments, supporting the notion of avoiding randomly initialized weights and instead using continual learning techniques to solve complex tasks. 

Source code for both the training methods used in this paper as well as the \textit{Dex} toolkit source code can be found at \href{https://github.com/innixma/dex}{\texttt{github.com/innixma/dex}}.

\subsubsection*{Acknowledgments}

We thank Vittorio Romeo for designing \textit{Open Hexagon}.
\bibliographystyle{abbrvnat}



\end{document}